\pdfoutput=1

\documentclass[11pt]{article}

\usepackage[final]{acl}

\usepackage{times}
\usepackage{latexsym}

\usepackage[T1]{fontenc}

\usepackage[utf8]{inputenc}

\usepackage{microtype}

\usepackage{inconsolata}

\usepackage{booktabs}
\usepackage{amssymb}
\usepackage{pifont}
\usepackage{tikz}
\usetikzlibrary{positioning} 
\usepackage{xcolor} 
\usetikzlibrary{shapes.geometric}
\usepackage{url}
\usepackage{tikz}
\usetikzlibrary{backgrounds,mindmap}
\usetikzlibrary{calc,positioning,intersections}
\usepackage{pgfplots}
\usepackage{listings}
\usepackage{multirow}
\usepackage[edges]{forest}
\usepackage{amsmath}
\newcommand{\cmark}{\ding{51}}
\newcommand{\xmark}{\ding{55}}

\definecolor{lightcoral}{rgb}{0.94, 0.5, 0.5}
\definecolor{lightgreen}{rgb}{0.56, 0.93, 0.56}
\definecolor{harvestgold}{rgb}{0.98, 0.85, 0.40}
\definecolor{brightlavender}{rgb}{0.75, 0.58, 0.89}
\definecolor{capri}{rgb}{0.0, 0.75, 1.0}
\definecolor{carminepink}{rgb}{0.92, 0.3, 0.26}
\definecolor{celadon}{rgb}{0.67, 0.88, 0.69}
\definecolor{darkpastelgreen}{rgb}{0.01, 0.75, 0.24}

\definecolor{hidden-draw}{RGB}{205, 44, 36}
\definecolor{hidden-blue}{RGB}{194,232,247}
\definecolor{hidden-orange}{RGB}{243,202,120}
\definecolor{hidden-yellow}{RGB}{242,244,193}
\definecolor{tree-level-1}{RGB}{245,20,85}
\definecolor{tree-level-2}{RGB}{246,86,118}
\definecolor{tree-level-3}{RGB}{248,177,193}
\definecolor{tree-leaf}{RGB}{176,230,198}

\definecolor{Self}{RGB}{255,0,128}
\definecolor{Ensemble}{RGB}{0,127,255}
\definecolor{Iterative}{RGB}{153,51,255}

\definecolor{exemplar1}{RGB}{136,98,148}
\definecolor{exemplar2}{RGB}{148,210,242}
\definecolor{knowledge1}{RGB}{249,219,152}
\definecolor{knowledge2}{RGB}{255,245,220}

\tikzstyle{my-box}=[
    rectangle,
    draw=hidden-draw,
    rounded corners,
    text opacity=1,
    minimum height=1.5em,
    minimum width=5em,
    inner sep=2pt,
    align=center,
    fill opacity=.5,
]
\tikzstyle{cause_leaf}=[my-box, minimum height=1.5em,
    fill=harvestgold!20, text=black, align=left,font=\scriptsize,
    inner xsep=2pt,
    inner ysep=4pt,
]
\tikzstyle{detect_leaf}=[my-box, minimum height=1.5em,
    fill=cyan!20, text=black, align=left,font=\scriptsize,
    inner xsep=2pt,
    inner ysep=4pt,
]
\tikzstyle{mitigate_leaf}=[my-box, minimum height=1.5em,
    fill=lightcoral!20, text=black, align=left,font=\scriptsize,
    inner xsep=2pt,
    inner ysep=4pt,
]
\tikzstyle{mitigate_leaf}=[my-box, minimum height=1.5em,
    fill=lightcoral!20, text=black, align=left,font=\scriptsize,
    inner xsep=2pt,
    inner ysep=4pt,
]
\tikzstyle{llm_leaf}=[my-box, minimum height=1.5em,
    fill=lightgreen!20, text=black, align=left,font=\scriptsize,
    inner xsep=2pt,
    inner ysep=4pt,
]
\tikzstyle{challenge_leaf}=[my-box, minimum height=1.5em,
    fill=brightlavender!20, text=black, align=left,font=\scriptsize,
    inner xsep=2pt,
    inner ysep=4pt,
]

\title{Medical Dialogue: A Survey of Categories, Methods, Evaluation and Challenges}

\author{
 Xiaoming Shi\textsuperscript{\rm 1}, Zeming Liu\textsuperscript{\rm 2}\thanks{\quad Corresponding author.}, Li Du\textsuperscript{3}, Yuxuan Wang\textsuperscript{4}, Hongru Wang\textsuperscript{5}, \\
 \textbf{Yuhang Guo\textsuperscript{6}, Tong Ruan\textsuperscript{7}, Jie Xu\textsuperscript{1}, Shaoting Zhang\textsuperscript{1}\footnotemark[1]}\\
 \textsuperscript{1} Shanghai AI Lab, Shanghai, China \textsuperscript{2} Beihang University, Beijing, China \\
 \textsuperscript{3} Beijing Academy of AI, Beijing, China \textsuperscript{4} Zhijiang Lab, Hangzhou, China \\
 \textsuperscript{5} The Chinese University of Hong Kong, Hong Kong \textsuperscript{6} Beijing Institute of Technology, Beijing, China \\
 \textsuperscript{7} East China University of Science and Technology, Shanghai, China \\
 {\tt \{shixiaoming, shaotingzhang\}@pjlab.org.cn; zmliu@buaa.edu.cn} \\ 
}

\begin{document}
\maketitle
\begin{abstract}
This paper surveys and organizes research works on medical dialog systems, which is an important yet challenging task.
Although these systems have been surveyed in the medical community from an application perspective, a systematic review from a rigorous technical perspective has to date remained noticeably absent.
As a result, an overview of the categories, methods, and evaluation of medical dialogue systems remain limited and underspecified, hindering the further improvement of this area.
To fill this gap,  we investigate an initial pool of 325 papers from well-known computer science, and natural language processing conferences and journals, and make an overview.
Recently, large language models have shown strong model capacity on downstream tasks, which also reshaped medical dialog systems' foundation.
Despite the alluring practical application value, current medical dialogue systems still suffer from problems.
To this end, this paper lists the grand challenges of medical dialog systems, especially of large language models.
\end{abstract}

\begin{figure*}[t]
    \centering
    \resizebox{\textwidth}{!}{
        \begin{forest}
            forked edges,
            for tree={
                grow=east,
                reversed=true,
                anchor=base west,
                parent anchor=east,
                child anchor=west,
                base=left,
                font=\small,
                rectangle,
                draw=hidden-draw,
                rounded corners,
                align=left,
                minimum width=4em,
                edge+={darkgray, line width=1pt},
                s sep=3pt,
                inner xsep=2pt,
                inner ysep=3pt,
                ver/.style={rotate=90, child anchor=north, parent anchor=south, anchor=center},
            },
            where level=1{text width=7.0em,font=\scriptsize,}{},
            where level=2{text width=5.5em,font=\scriptsize,}{},
            where level=3{text width=7.2em,font=\scriptsize,}{},
            where level=4{text width=6.4em,font=\scriptsize,}{},
            [
                Medical Dialog System, ver, color=carminepink!100, fill=carminepink!15, text=black
                [
                    System Categories (\S \ref{sec:background}), color=harvestgold!100, fill=harvestgold!100, text=black
                    [                    
                        Functions, color=harvestgold!100, fill=harvestgold!60,  text=black
                        [
                        Diagnosis, color=harvestgold!60, fill=harvestgold!20,  text=black
                            [{\, KS-DS~\cite{xu2019end}, KNSE~\cite{chen-etal-2023-knse}, DISC-MedLLM~\cite{bao2023disc}, BenTsao~\cite{wang2023huatuo}}, cause_leaf, text width=31.5em]
                        ]
                        [
                        Intervention, color=harvestgold!60, fill=harvestgold!20,  text=black
                            [{\, TRIK~\cite{ljunglof2009trik}, Lekbot~\cite{ljunglof2011lekbot}, ASD~\cite{ali2020virtual,di2020explorative}}, cause_leaf, text width=31.5em]
                        ]
                        [
                        Monitoring, color=harvestgold!60, fill=harvestgold!20,  text=black
                            [{\, Hear me out~\cite{maharjan2019hear}, Dr. Youth~\cite{lee2019development}, I hear you I feel you~\cite{lee2020hear}}, cause_leaf, text width=31.5em]
                        ]
                        [
                        Counselling, color=harvestgold!60, fill=harvestgold!20,  text=black
                            [{\, Psychiatric Counseling~\cite{oh2017chatbot}, Robo~\cite{moghadasi2020robo}, Woebot~\cite{de2020investigating}}, cause_leaf, text width=31.5em]
                        ]
                        [
                        Education, color=harvestgold!60, fill=harvestgold!20,  text=black
                            [{\, VirtualPatient~\cite{campillos2020designing}, SOPHIE~\cite{ali2021novel}}, cause_leaf, text width=31.5em]
                        ]
                        [
                        Multi-objective, color=harvestgold!60, fill=harvestgold!20,  text=black
                            [{\, Robo~\cite{moghadasi2020robo}, Woebot~\cite{de2020investigating}, Therapy Chatbot~\cite{sharma2018digital}}, cause_leaf, text width=31.5em]
                        ]
                    ]                    
                    [
                        Types, color=harvestgold!100, fill=harvestgold!60, text=black
                        [
                        Task-oriented Dialogue, color=harvestgold!60, fill=harvestgold!20,  text=black
                            [{\, TeenChat~\cite{huang2015teenchat}, DQN-Agent~\cite{liu2018dialogue}, KS-DS~\cite{xu2019end}, HRL~\cite{liao2020task}}, cause_leaf, text width=31.5em]
                        ]                    
                       [
                        Dialogue Recommendation, color=harvestgold!60, fill=harvestgold!20,  text=black
                            [{\, CF~\cite{huang2012collaboration}, SPA-ACA~\cite{hoens2013reliable}, fb-kNN~\cite{bhatti2019recommendation}, DP-CRNN~\cite{zhou2020cnn}}, cause_leaf, text width=31.5em]
                        ]
                        [
                        ChitChat Dialogue, color=harvestgold!60, fill=harvestgold!20,  text=black
                            [{\, EmotionalChat  ~\cite{huber2018emotional,lan2021chinese,chen2022wish,zhang2023refashioning,zhao2023chatgpt}}, cause_leaf, text width=31.5em]    
                        ]
                        [
                        Question-and-answering , color=harvestgold!60, fill=harvestgold!20,  text=black
                            [{\, MEANS~\cite{abacha2015means}, BenTsao~\cite{wang2023huatuo}, DISC-MedLLM~\cite{bao2023disc}}
                            , cause_leaf, text width=31.5em]
                        ]
                        [
                        Mixed-type Dialogue, color=harvestgold!60, fill=harvestgold!20,  text=black
                            [{\, InsMed~\cite{shi-etal-2023-midmed}}, cause_leaf, text width=31.5em]
                        ]
                    ]
                ]
                [
                    Methods before LLM (\S \ref{sec:traditional}),
                    color=cyan!100, fill=cyan!100, text=black
                    [   
                        Retrieval, color=cyan!100, fill=cyan!60, text=black
                        [
                            Literature Retrieval, color=cyan!100, fill=cyan!30, text=black
                            [
                                {\, SemBioNLQA~\cite{sarrouti2020sembionlqa}, BioMedBERT~\cite{chakraborty-etal-2020-biomedbert}, MedCPT~\cite{jin2023medcpt}}
                                , detect_leaf, text width=31.5em
                            ]
                        ]
                        [
                            Dialogue Retrieval, color=cyan!100, fill=cyan!30, text=black
                            [
                                {\, SHIHbot~\cite{brixey2017shihbot}, Healthcare Bot~\cite{athota2020chatbot}}, detect_leaf, text width=31.5em
                            ]
                        ]
                    ]
                    [
                        Generation, color=cyan!100, fill=cyan!60, text=black
                        [
                            Pipeline, color=cyan!60, fill=cyan!30, text=black
                            [
                                Natural Language \\ Understanding, color=cyan!30, fill=cyan!10, text=black
                                [
                                    {\,~Token-level: ULisboa~\cite{leal2015ulisboa}, BERT~\cite{miftahutdinov2019deep}, \\ \quad \quad \quad \quad MTAAL~\cite{zhou2021mtaal}, generate-and-rank~\cite{xu2020generate} \\
                                    ~Utterance-level: \\ \quad \quad \quad \quad BiGRU~\cite{li2019joint}, MSL~\cite{shi2021understanding}, MedDG~\cite{liu2022meddg} \\
                                    ~Dialog-level: SAT~\cite{du2019extracting}, MIE~\cite{zhang2020mie}, \\ \quad \quad \quad \quad CMUI~\cite{dai2022chinese}, CSDM~\cite{zeng2022csdm}}, detect_leaf, text width=23.5em
                                ]
                            ]
                            [
                                Dialogue Management, color=cyan!30, fill=cyan!10, text=black
                                [
                                    {\, DQN Agent~\cite{wei2018task}, KR-DQN~\cite{xu2019end}, \\ DSMD~\cite{liu2022my}, HRL~\cite{zhong2022hierarchical}}
                                    , detect_leaf, text width=23.5em
                                ]
                            ]
                            [
                                Natural Language \\ Generation, color=cyan!30, fill=cyan!10, text=black
                                [
                                    {\, VRBot~\cite{li2021semi},GEML~\cite{lin2021graph},KnowInject~\cite{naseem2022incorporating}}
                                    , detect_leaf, text width=23.5em
                                ]
                            ]
                        ]
                        [
                            End-to-end, color=cyan!60, fill=cyan!30, text=black
                            [
                                    {\, MedDialog~\cite{zeng-etal-2020-meddialog}, VA~\cite{saha2021large}, CovidDialog~\cite{zhou2021generation}, MedPIR~\cite{zhao2022medical}}, detect_leaf, text width=31.5em
                                ]
                        ]   
                    ]
                    [
                        Hybrid, color=cyan!100, fill=cyan!60, text=black
                        [
                            {\, HRGR-Agent~\cite{li2018hybrid}, MedWriter~\cite{yang2021writing}, BIOREADER~\cite{frisoni2022bioreader}, PMC-Patients~\cite{zhao2022pmc}}, detect_leaf, text width=40.35em
                        ]
                    ]
                ]
                [
                LLM-based Methods (\S \ref{sec:llm}), color=lightcoral!100, fill=lightcoral!100, text=black
                        [
                            Prompting LLMs, color=lightcoral!60, fill=lightcoral!20, text=black
                            [
                            {\, DeID-GPT~\cite{liu2023deid}, ChatCAD~\cite{wang2023chatcad}, Dr. Knows~\cite{gao2023leveraging}, MedPrompt~\cite{nori2023can}, \\ MedPaLM~\cite{singhal2023large}, MedPaLM2~\cite{singhal2023towards}, MedAgent~\cite{tang2023medagents}}, mitigate_leaf, text width=40.35em
                            ]
                        ]
                        [
                            Fine-tuning LLMs, color=lightcoral!60, fill=lightcoral!20, text=black
                            [
                                {\, PULSE~\cite{pulse2023}, BenTsao~\cite{wang2023huatuo}, HuatuoGPT~\cite{zhang2023huatuogpt}, ChatDoctor~\cite{yunxiang2023chatdoctor}, \\ DoctorGLM~\cite{xiong2023doctorglm}, Zhongjing~\cite{yang2023zhongjing}, Qilin-med~\cite{ye2023qilin}, AMIE~\cite{tu2024towards}, \\ MEDITRON~\cite{chen2023meditron}, Radiology-LLaMA2~\cite{liu2023radiology}, Clinical Camel~\cite{toma2023clinical}, XrayGLM~\cite{wang2023XrayGLM}}, mitigate_leaf, text width=40.35em
                            ]
                        ]
                ]
                [
                    Evaluation (\S \ref{sec:evaluation}), color=lightgreen!100, fill=lightgreen!100, text=black
                    [
                        Evaluation Metrics, color=lightgreen!100, fill=lightgreen!60, text=black
                        [
                            {\, Retrieval: Mean Average Precision~\cite{luo2022improving}, etc. \\ 
                            \, Pipeline: Precision, Recall, F1, Accuracy, etc.~\cite{qin-etal-2023-end} \\
                            \, End-to-end Generation: BLUE~\cite{papineni2002bleu}, ROUGE~\cite{lin2004rouge}, Distinct~\cite{li2015diversity}, human evaluation~\cite{shi-etal-2023-midmed}, etc. 
                            }, llm_leaf, text width=40.35em
                        ]
                    ]
                    [
                        Datasets, color=lightgreen!100, fill=lightgreen!60, text=black
                        [
                            {\, Retrieval: BioASQ~\cite{luo2022improving}\\
                             \, Pipeline: CMDD~\cite{lin2019enhancing}, MIE~\cite{zhang2020mie}, MedDG~\cite{liu2022meddg}, IMCS-21~\cite{chen2022benchmark}, \\ \quad \quad \quad \quad \quad \quad   MZ~\cite{wei2018task}, DX~\cite{xu2019end}\\
                             \, End-to-end Generation: MedDialog~\cite{zeng2020meddialog}, MedDG~\cite{liu2022meddg}, MidMed~\cite{shi-etal-2023-midmed}, \\ \quad \quad \quad \quad \quad \quad CovidDialog~\cite{yang2020generation}, Ext-CovidDialog~\cite{varshney2023knowledge}
                            },  llm_leaf, text width=40.35em
                        ]
                    ]
                ]
                [
                    Grand Challenges (\S \ref{sec:challenge}), color=brightlavender!100, fill=brightlavender!100, text=black
                    [
                        Challenges Inherited from General Domain, color=brightlavender!100, fill=brightlavender!60, text=black, text width=12em
                        [
                        {\, Hallucination~\cite{huang2023survey}, Numberical Data~\cite{akhtar2023exploring}, Adversarial Attack~\cite{shayegani2023survey}}, challenge_leaf, text width=33.9em
                        ]    
                    ]
                    [
                        Medical-specific Challenges, color=brightlavender!100, fill=brightlavender!60, text=black, text=black, text width=12em
                        [
                            {\, Medical Specialization~\cite{gao2023leveraging}, Medical Evaluation~\cite{cai2023medbench}, \\ \, Multi-modal Dialogue~\cite{wang2023chatcad}, Multi-disciplinary Treatment~\cite{tang2023medagents}}, challenge_leaf, text width=33.9em
                        ]
                    ]
                ]
        ]
        \end{forest}}
    \caption{The main content flow and categorization of this survey.}
    \label{figure:categorization_of_survey}
\end{figure*}
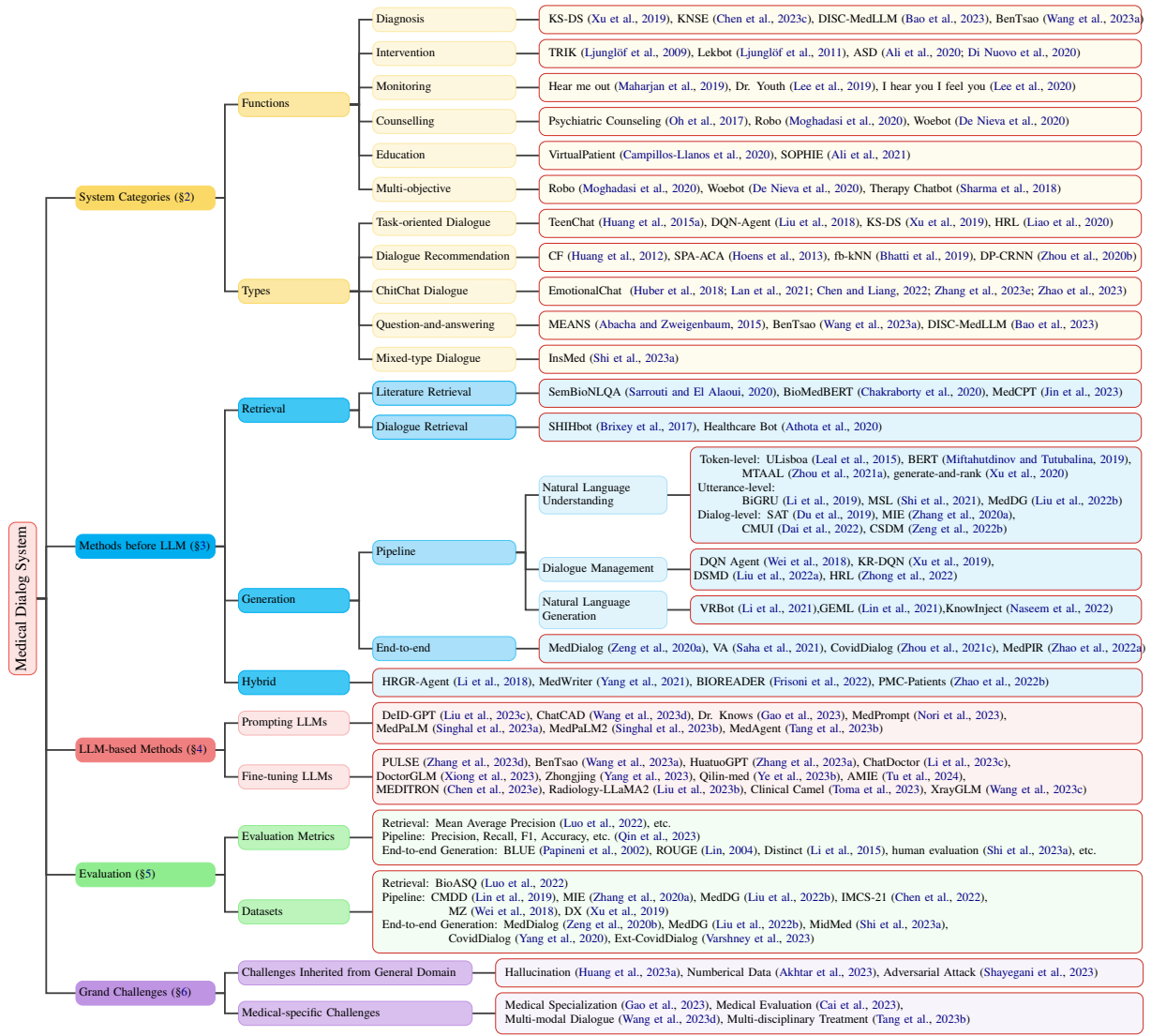

\section{Introduction}
Dialogue systems for the medical domain, which are designed to converse with patients to obtain additional symptoms, make a diagnosis and recommend a treatment plan automatically~\cite{tang2016inquire,wei2018task,liao2020task,zhong2022hierarchical}. 
Medical dialogue systems have significant potential to simplify the diagnostic procedure and reduce the cost of collecting information from patients, thus containing alluring application value and attracting academic and industrial attention~\cite{wang2023huatuo,chen2023bianque}.

Existing medical dialogue systems have played an important role in diagnosis~\cite{liao2020task,lin2019enhancing}, monitoring~\cite{lee2019development,maharjan2019hear}, intervention~\cite{javed2018robot}, counselling~\cite{lee2017chatbot}, 
education~\cite{ali2021novel}, and etc.
To meet these real meets, retrieval~\cite{tao2021building,zhu2022knowledge}, generation~\cite{zhong2022hierarchical,liu2022meddg,du2019extracting}, and hybrid~\cite{li2018hybrid,yang2021writing} methods are applied for building medical dialogue systems.
Specifically, retrieval-based methods select appropriate responses from a pre-built index,
generation-based methods respond in a generative manner,
and hybrid methods combine both approaches, using retrieval for efficiency and generative methods for flexibility.

Recently, the revolutionary progress in \textbf{l}arge \textbf{l}anguage \textbf{m}odels (LLM)~\cite{zeng2022glm,openai2023gpt4,touvron2023llama,bao2023disc} has catalyzed substantial technological transformations in dialogue systems.
LLMs are sophisticated neural network-based systems that have been trained on vast amounts of text data, enabling them to generate human-like responses and achieve remarkable accuracy, thus reshaping medical dialogue systems' foundation.

Despite the potential performance in the medical question-answering, 
there remains a translational gap~\cite{newman2021translational}\footnote{Translational NLP research is focused on identifying the factors that contribute to the success or failure of translations and on creating versatile and adaptable methodologies that can bridge the gap between theoretical NLP advancements and their practical implementation in various real-world scenarios.} between cutting-edge techniques and realistic requirements in various medical scenarios.
For example, in Figure~\ref{figure:example}, LLMs operate in a question-and-answering manner, instead of diagnosing like doctors, 
which may lead to patients being unable to obtain precise diagnostic results and effective treatment strategies.

To move towards closing this gap, this work 
(a) summarizes the system categories, methods, and evaluation of medical dialogue systems, 
(b) analyzes the current issues and challenges of medical dialogue systems, then 
(c) attempts to provide potential solutions to facilitate further development.

In the existing literature, medical dialogue systems have been discussed and surveyed~\cite{laranjo2018conversational,vaidyam2019chatbots,kearns2019systematic,valizadeh2022ai,he2023survey,hadi2023survey}.
This survey differs from these surveys in two aspects.
First, our survey is a systematic review from a rigorous technical perspective, summarizing methods before LLM, and LLM-based methods.
Second, this survey highlights the grand challenges of current medical dialogue systems, including medical specialization~\cite{gao2023leveraging} and multi-disciplinary treatment~\cite{tang2023medagents}, which may inspire further research.

This survey is organized as follows:
Section~\ref{sec:background} introduces the system categories of medical dialogue systems,
followed by the methods before LLM, LLM-based methods, and evaluation of medical dialogue systems in Section~\ref{sec:traditional}, Section~\ref{sec:llm}, and Section~\ref{sec:evaluation}, respectively.
Finally, we summarize the major challenges and possible solutions for further work in Section~\ref{sec:challenge}.

The contributions of this paper are as follows:
\begin{itemize}
    \item \textbf{\textit{First survey:}} To our knowledge, we are the first to present a comprehensive survey for medical dialogue systems from a technical perspective, summarizing categories, methods before LLM, and LLM-based methods.  
    \item \textbf{\textit{New frontiers:}} We discuss frontiers and summarize common and medical-specific challenges, which shed light on further research;
    \item \textbf{\textit{Abundant resources:}} 
    We make the first attempt to organize medical dialogue resources including open-source implementations, corpora, and paper lists, which may help new researchers quickly adapt to this field.
\end{itemize}

\section{System Categories}
\label{sec:background}
This section briefly summarizes the functions and techniques of current medical dialogue systems.
Specifically, the functions can be divided into five categories according to the dominant subjects, including doctors (diagnosis, intervention), patients (monitoring, counseling), and medical students (medical education).

\subsection{Functions of Systems}
It is essential to figure out the different functions of medical dialogues in our daily life.
As shown in Figure~\ref{figure:categorization_of_survey}, there are eight main system objectives: 

\noindent
\textbf{Diagnosis systems}~\cite{xu2019end,chen-etal-2023-knse,bao2023disc,wang2023huatuo} are designed to first collect the patient's medical history, symptoms, signs, and then predict health condition;

\noindent
\textbf{Intervention systems}~\cite{ljunglof2009trik,ljunglof2011lekbot,ali2020virtual,di2020explorative} are designed to provide comprehensive approaches and strategies to prevent diseases, cure or reduce the severity or duration of diseases;

\noindent
\textbf{Monitoring systems}~\cite{maharjan2019hear,lee2019development,lee2020hear} are designed to continuously track, record, and analyze vital signs and other health-related data of patients;

\noindent
\textbf{Counseling systems}~\cite{oh2017chatbot,moghadasi2020robo,de2020investigating} are designed to guide medical counseling services, such as recommending hospitals and doctors;

\noindent
\textbf{Medical education systems}~\cite{ali2021novel} are designed to provide a simulation of real clinical scenarios. A typical application is patient simulator systems~\cite{sijstermans2007training,danforth2009development,menendez2015using};

\noindent
\textbf{Multi-objective systems}~\cite{moghadasi2020robo,de2020investigating,sharma2018digital} are designed for more than one of those goals.

The above are brief descriptions of the different functions of medical dialogue systems,
and new functions of medical dialogue systems will emerge according to emergent user demands,
which gives rise to various challenges.

\subsection{Types of Dialogues}
From the aspect of the dialogue type, current medical dialogue systems can be divided into five categories:

\noindent
\textbf{Task-oriented dialogue systems}~\cite{young2013pomdp,huang2015teenchat,liu2018dialogue,xu2019end,liao2020task} are designed to help users complete specific tasks through dialogue interaction;

\noindent
\textbf{Dialogue recommendation systems}~\cite{huang2012collaboration,hoens2013reliable,bhatti2019recommendation,zhou2020detecting,ko2022survey} are designed to recommend information, products, or services that users may be interested in by analyzing users' historical behavioral data and portraits;

\noindent
\textbf{Chit-chat dialogue systems}~\cite{huber2018emotional,lan2021chinese,chen2022wish,yan2022deep,zhang2023refashioning,zhao2023chatgpt} are designed to revolve around exchanging information and discussing topics with users;

\noindent
\textbf{Question-and-answer systems}~\cite{abacha2015means,zaib2022conversational,wang2023huatuo,bao2023disc} are designed to provide relevant and accurate answers according to users' specific questions;
    
\noindent
\textbf{Mixed-type dialogue systems}~\cite{shi-etal-2023-midmed} are designed to finish complex tasks by the combination of the above four types of dialogues.

\section{Methods before LLM}
\label{sec:traditional}
In this section, we briefly summarize the medical dialogue systems before the emergence of LLM from the technical aspect of publicly available resources.
The methods can be mainly divided into three categories, retrieval-based methods, generation-based methods, and hybrid methods~\cite{wang2023survey}.

\subsection{Retrieval-based methods}
Retrieval-based medical dialogue systems are designed to select appropriate responses from the pre-built index~\cite{tao2021building,zhu2022knowledge}, 
which can be mainly divided into two categories according to different sources of indexed documents, medical literature, and medical dialogue.

\noindent
\textbf{Medical Literature Retrieval}.
The recent statistics show that 61\% of adults look online for health information~\cite{fox2011social}.
This demands proper retrieval systems for health-related biomedical queries.
Major challenges in the biomedical domain are in handling complex, ambiguous, inconsistent medical terms and their ad-hoc abbreviations~\cite{zhao2019neural,luo2019mcn}.

Biomedical information retrieval has traditionally relied upon term-matching algorithms (such as TF-IDF~\cite{aizawa2003information}, BM25~\cite{robertson1995okapi}, and In\_expC2~\cite{robertson2009probabilistic,sankhavara2018biomedical}), which search for documents that contain terms mentioned in the query.

However, term-matching suffers from semantic retrieval, especially for terms that have different meanings in different contexts.
To alleviate the issue, \citet{luo2022improving} provides context-specific vector representations for each context and query, and the matching is conducted with the vector similarity.
Vector representation focuses on semantic information, thus alleviating semantic inconsistency.

\noindent
\textbf{Medical Dialogue Retrieval}.
Medical dialogue systems with the dialogue retrieval method are designed to select appropriate responses from the pre-built dialogue index.
Typically, medical dialogue retrieval methods choose responses that are ranked highest but may choose a lower-ranked response to avoid repetition~\cite{athota2020chatbot,brixey2017shihbot}.
The selection is conducted with a response classifier that is trained on linked questions and responses.
If the score of the top-ranked response is below a predefined threshold, the medical dialog systems instead select an off-topic response that indicates ``I do not understand''.

Despite the efficiency, the results may not exactly match patients' queries, which may trigger serious safety risks.


\subsection{Generation-based Methods}
Generation-based methods can be divided into two categories, pipeline and end-to-end.
The pipeline methods typically generate system response through sub-components, 
while the end-to-end methods directly generate system response given only dialogue history and the corresponding knowledgebase without intermediate supervision.

\subsubsection{Pipeline}
The pipeline methods mainly contain three sub-components (natural language understanding, dialogue management, and natural language generation)~\cite{young2013pomdp}.

\noindent
\textbf{Natural Language Understanding}.
Natural language understanding for medical dialogue is designed to capture key semantic meaning~\cite{zhang2020recent}.
This work divides the medical natural language understanding task into three levels: token-level (medical concept normalization), utterance-level (slot filling, intent detection), and dialogue-level (medical dialogue information extraction).



\noindent
\textit{Token-level}.
Medical concept normalization aims to map a variable length medical mention to a medical concept in some external coding system.
The technique development can be summarized as: string-matching or dictionary look-up approach~\cite{leal2015ulisboa,d2015sieve,lee2016audis},
deep learning based classification method~\cite{limsopatham2016normalising,miftahutdinov2019deep,luo2018multi,zhao2019neural,zhou2021end,zhou2021mtaal},
generate-and-rank method~\cite{xu2020generate}, constrained generation method~\cite{yan2020knowledge}.

\noindent
\textit{Utterance-level}.
Intent detection and slot filling are utterance-level natural language understanding tasks.
An intent specifies the goal underlying the expressed utterance while slots are additional parameters for these intents.
Intent detection is usually defined as a multi-label classification problem and slot filling is usually defined as a sequence labeling problem~\cite{weld2022survey,liu2022meddg}.
To further utilize semantic information from these two tasks, intent detection and slot filling are usually jointly learned~\cite{zhang-etal-2019-joint,li2019joint,song-etal-2022-enhancing}.

\noindent
\textit{Dialog-level}.
Medical dialogue information extraction is designed to extract key information from medical dialogues, 
which greatly facilitates the development of many real-world applications such as electronic medical record generation~\cite{guan2018generation}, automatic disease diagnosis~\cite{xu2019end}, etc.
\citet{du2019extracting,zhang2020mie} propose to convert doctor-patient dialogues into electronic medical records, effectively reducing the labor costs of doctors.
To enhance the exploitation of the inter-dependencies in multiple utterances, \citet{dai2022chinese} introduces a selective attention mechanism to explicitly capture the dependencies among utterances.
Furthermore, to alleviate the issue of speaker role ambiguity, \citet{zeng2022csdm} introduces a multi-view aware channel that captures different information in dialogues.


\noindent
\textbf{Dialog Management}.
Dialog management aims to select the next actions for response based on the current dialog state toward achieving long-term dialog goals~\cite{young2013pomdp,thrun2000reinforcement,schatzmann2006survey}.

\citet{wei2018task,xu2019end} cast the medical dialogue system as a Markov Decision Process and train the dialogue policy via reinforcement learning, which is composed of states, actions, rewards, policy, and transitions.
Besides, \citet{zhong2022hierarchical} propose to integrate a hierarchical policy structure of two levels into the dialog system for policy learning, alleviating the huge action space in the real environment.
In addition, \citet{liu2022my} propose an interpretable decision process to enhance interpretability.


\noindent
\textbf{Natural Language Generation}.
Natural language generation is designed to convert system acts into text or speech~\cite{young2013pomdp}.
\citet{li2021semi} propose to summarize diagnosis history through a key phrase and propose a variational Bayesian generative approach to generate based on patient states and physician actions.
Besides, to enhance the rationality of medical dialogues, \citet{naseem2022incorporating} leverage an external medical knowledge graph and injects triples as domain knowledge into the dialogue generation.
To capture the correlations between different diseases, \citet{lin2021graph} propose to utilize a commonsense knowledge graph to characterize the prior disease-symptom relations.

\subsubsection{End-to-end Dialogue Generation}
End-to-end dialogue generation aims to directly generate responses based on dialogue history and knowledgebase, which mostly adopts a sequence-to-sequence framework~\cite{bahdanau2014neural,vaswani2017attention}.
It consists of a context encoder to encode the dialogue history and a decoder to generate the responses.
Formally, give a sequence of inputs $(x_{1}, \dots, x_{T})$, 
the goal of the task is to estimate the conditional probability $p(y_{1}, \dots, y_{T^{'}} | x_{1}, \dots, x_{T})$, 
where $(y_{1}, \dots, y_{T^{'}})$ is the output sequence, $T$ is the input sequence length and $T^{'}$ is the output sequence length.
The  probability of $y_{1}, \dots, y_{T^{'}}$ is usually computed in the autoregressive manner. 
Each $p(y_{t} | x_{1}, \dots, x_{T}, y_{1}, \dots, y_{t-1})$ distribution is represented with a softmax overall words.

In the medical domain, \citet{saha2021large}, \citet{yang2020generation,zhou2021generation} and \citet{zeng2020meddialog} apply the above generative models for dialogues on mental health, COVID-19 and diagnosis, respectively.
Besides, \citet{zhao2022medical} build a medical dialogue graph that exploits the medical relationship between utterances and trains the model to generate the pivotal information before producing the actual response, thus learning to focus on the key information.



\subsection{Hybrid Methods}
Due to the limited coverage and timeliness of training data, generation-based models often result in hallucination~\cite{ji2023survey,ye2023cognitive}, which are particularly severe in medical scenarios and may lead to serious risks.
To alleviate the issue, retrieval augmented generation methods~\cite{lewis2020retrieval,li2022survey} are proposed, which retrieve accurate and in-time information to augment generation to obtain precise responses.

In the medical domain, BIOREADER~\cite{frisoni2022bioreader} fetches and assembles relevant scientific literature chunks from a neural database, and then enhances the domain-specific T5-based solution~\cite{2020t5}.
By contrast, \citet{zhao2022pmc} retrieve PubMed Central articles using simple heuristics and Retrieved articles are utilized as supplementary materials for generating responses for clinical decision-supporting systems.
For report generation, MedWriter~\cite{yang2021writing} first employs the retrieval module to retrieve the most relevant sentences from retrieved reports for given images,
and then fuses them to generate meaningful medical reports.
For sentence retrieval, HRGR-Agent~\cite{li2018hybrid} utilizes reinforcement learning with sentence-level and word-level rewards.

\section{Large Language Model-based Methods} 
\label{sec:llm}
LLMs have generated significant interest due to their remarkable performance in understanding instructions and generating human-like responses.
This section summarizes the medical dialogue methods based on LLMs.
Current LLMs-based methods can be divided into two categories, prompting, and fine-tuning general LLMs.
Typical medical LLMs are listed in Table~\ref{tbl:llm_sum}.

\subsection{Prompting based Methods}
The training corpora of LLMs contain medical literature, therefore it is possible to align the LLMs with medical scenarios through appropriate prompts.
Popular prompting methods include hand-crafted prompting and prompt tuning.

\subsubsection{Hand-crafted Prompting}
Hand-crafted prompting is designed to create intuitive prompts based on human introspection.
Specifically, hand-crafted prompting methods in the current medical dialogue system can be mainly divided into three categories, zero/few-shot prompting, chain-of-thought prompting, and prompting ensemble.

\noindent
\textbf{Zero/Few-shot Prompting}.
Zero-shot prompting aims to directly give instructions to prompt LLMs to efficiently perform a task following the given instruction.
Meanwhile, the few-shot prompting strategy aims to include samples describing the task through demonstrations, which has shown effectiveness in various tasks~\cite{brown2020language,min2022rethinking}.

In the medical domain, expert hand-crafted prompting is widely utilized.
Current works mainly focus on question-and-answering~\cite{nori2023can,singhal2023large,singhal2023towards}, diagnosis~\cite{wang2023chatcad,gao2023leveraging,tang2023medagents}, and text de-identification~\cite{liu2023deid}.
For question-and-answering, MedPaLM~\cite{singhal2023large}, MedPaLM 2~\cite{singhal2023towards}, and MedPrompt~\cite{nori2023can} collaborate with a panel of qualified clinicians to identify the best demonstration examples and meticulously craft the few-shot prompts.
For diagnosis, ChatCAD~\cite{wang2023chatcad}, Dr. Knows~\cite{gao2023leveraging}, MedAgents~\cite{tang2023medagents} design task-specific prompts for computer-aided diagnosis on medical image, diagnosis prediction, and multi-disciplinary treatment, respectively.
Another LLM prompting applied in the medical domain is text de-identification and anonymization of medical reports~\cite{liu2023deid}.

\noindent
\textbf{Chain-of-Thought Prompting}.
\textbf{C}hain-\textbf{o}f-\textbf{T}hought (CoT) improves LLMs' ability to solve complex problems by encouraging it to explain its reasoning process step by step before generating answers~\cite{wei2022chain}. 

In the medical domain, medical questions involve complex multi-step reasoning, making them a good fit for CoT prompting techniques. 
\citet{singhal2023large,singhal2023towards} craft CoT prompts to provide clear demonstrations of how to reason and answer the given medical questions.
Besides, MedPrompt~\cite{nori2023can} utilizes GPT-4 to generate CoT with task-specific prompts and to mitigate the hallucinated or incorrect reasoning chains, MedPrompt~\cite{nori2023can} utilizes the label-verification.
Specifically, GPT-4 is required to generate both a rationale and an estimation of the most likely answer to follow from that reasoning chain, and the reliability of generated chains is judged by whether the answers match the ground truth label.

\noindent
\textbf{Self-consistency Prompting}.
MedPaLM~\cite{singhal2023large} and MedPaLM2~\cite{singhal2023towards} utilize self-consistency prompting~\cite{wang2022self} to improve the performance on the multiple-choice benchmarks by prompt and sample multiple decoding outputs from the model.
The method is based on the rationale that for the medical domain with complex reasoning paths, there might be multiple potential routes to the correct answer~\cite{singhal2023large}.
Therefore, the final answer is the one with the majority vote.

\subsubsection{Prompt Tuning}
The above methods utilize hand-crafted static prompts, which are knowledge-intensive and training-free.
To better align general LLMs with the medical domain, 
inspired by the great success of prompting~\cite{liu2023pre} and fine-tuning~\cite{hu2023llm}, 
prompt tuning~\cite{liu2021p,lester2021power} introduces learnable prompts, 
which is fine-tuned during the training stage.
In contrast to traditional fine-tuning methods~\cite{hu2023llm}, 
prompt tuning only fine-tuning a very small set of parameters, 
thus effectively aligning LLMs to the medical domain~\cite{nori2023can}.
Recently, MedPaLM~\cite{singhal2023large} and MedPaLM2~\cite{singhal2023towards} apply the prompt tuning in various medical question-and-answer datasets and achieve a competitive performance compared to human experts.

\subsection{Fine-tuning Based Methods}

Compared to small-scale models, LLMs exhibit strong generalization across various natural language processing tasks and a unique emergent ability to solve unseen or complicated tasks. 
However, despite their numerous merits, LLMs are not designed to cater specifically to the medical domain. 
Their general domain knowledge often falls short when addressing such specialized fields, where accurate and domain-specific expert knowledge is critical. 
This can lead to sub-optimal diagnostic precision, drug recommendations, and medical advice, potentially endangering patients.
Recently, efforts have been made to address this problem.
The typical training method is fine-tuning foundation models on medical data.

Various works, including PULSE~\cite{pulse2023}, BenTsao~\cite{wang2023huatuo}, HuatuoGPT-II~\cite{chen2023huatuogpt}, ChatDoctor~\cite{yunxiang2023chatdoctor}, MEDITRON~\cite{chen2023meditron}, Radiology-LLaMA2~\cite{liu2023radiology}, Clinical Camel~\cite{toma2023clinical}, XrayGLM~\cite{wang2023XrayGLM}, conduct supervised fine-tuning by fine-tuning foundation models with their task-specific tasks.
Besides, Zhongjing~\cite{yang2023zhongjing} implements an entire training pipeline from continuous pre-training, and supervised fine-tuning, to reinforcement learning from human feedback,
and training with a Chinese multi-turn medical dialogue dataset, 
which enhances the model’s capability for complex dialogue.
In addition, to address the issue of overconfident predictions and tapping into domain-specific insights, Qilin-med~\cite{ye2023qilin} presents a method combining domain-specific continued pre-training, supervised fine-tuning, and direct preference optimization.

\section{Evaluation}
\label{sec:evaluation}

\subsection{Metrics}
In this section, we introduce two common evaluation methods: automatic and human evaluation.

\noindent
\textbf{Automatic Evaluation}.
For retrieval tasks, the mean average precision is utilized in \citet{luo2022improving}.
For the pipeline tasks (except dialogue policy learning), precision, recall, F1, and accuracy are utilized as the evaluation metrics~\cite{qin-etal-2023-end}.
For the end-to-end generation, automatic evaluation usually uses various indicators and evaluation tools, such as BLEU~\cite{papineni2002bleu}, ROUGE~\cite{lin2004rouge}, BERTScore~\cite{zhang2019bertscore}, etc., to quantify the similarity and quality between the model-returned results and the reference results.
Compared with human evaluation, automatic evaluation does not require human participation, which saves costs and time.

\noindent
\textbf{Human Evaluation}.
Human evaluation is especially for generation tasks, due to the reason that free text of the generated model exhibits diverse expressions which are formally different, but semantically similar, thus is more reliable for generation tasks~\cite{novikova2017we}.
Compared with automatic evaluation, manual evaluation is closer to the actual application scenario and can provide more comprehensive and accurate feedback. 
In the manual evaluation, evaluators (such as experts, researchers, or ordinary users) are usually invited to evaluate generated results.
Despite the effectiveness, even human evaluations can have high variance and instability due to cultural and individual differences~\cite{peng1997validity}.

\subsection{Datasets and Benchmarks}
Medical evaluation datasets are used to test and compare the performance of different dialogue systems on various tasks.
We list fourteen popular datasets and benchmarks from Table~\ref{tbl:nlu_result} to Table~\ref{tbl:generation_result}.
Each benchmark focuses on different aspects and evaluation criteria, providing valuable contributions to their respective domains.
These benchmarks are divided into four categories.

\noindent
\textbf{Benchmarks for Retrieval}.
BioASQ~\cite{tsatsaronis2015overview} assesses the ability of systems to semantically index very large numbers of biomedical scientific articles and to return concise and user-understandable answers to given natural language questions by combining information from biomedical articles and ontologies.
The evaluation metric is the mean average precision.
The benchmark on BioASQ is listed in Table~\ref{tbl:retrieval_result}.

\noindent
\textbf{Benchmarks for Pipeline Tasks}.
Medical dialogue systems contain a vast majority of tasks.
To this end, existing benchmarks tend to evaluate the performance in different tasks.

For sentence-level natural language understanding, there are three datasets, including CMDD~\cite{lin2019enhancing}, MedDG~\cite{liu2022meddg}, and IMCS-21 (NER)~\cite{chen2022benchmark}.
The widely used evaluation metrics are precision, recall, and F1.
The benchmark is listed in Table~\ref{tbl:nlu_result}.
Besides, MIE~\cite{zhang2020mie} is a dialogue-level natural language understanding dataset.
The benchmark is listed in Table~\ref{tbl:MIE_result}.

For the dialogue act classification task, IMCS-21~\cite{chen2022benchmark} contains a sub-task.
The utilized evaluation metrics are precision, recall, F1, and accuracy.
The benchmark is listed in Table~\ref{tbl:dac_result.}.

For the dialogue policy learning, the benchmark on MZ~\cite{wei2018task}, DX~\cite{xu2019end}, IMCS-21 (DDP)~\cite{chen2022benchmark} is listed in Table~\ref{tbl:policy_results}.

\noindent
\textbf{Benchmarks for Generation Tasks}.
For the end-to-end dialogue generation, there are five dialogue datasets, including MedDialog~\cite{zeng-etal-2020-meddialog}, MedDG~\cite{liu2022meddg}, CovidDialog~\cite{yang2020generation}, Ext-CovidDialog~\cite{varshney2023knowledge}, and MidMed~\cite{shi-etal-2023-midmed}.
The automatic evaluation metrics utilized for the medical generation include BLEU-4, Distinct-1, Distinct-2, etc.
The benchmark is listed in Table~\ref{tbl:generation_result}.

\noindent
\textbf{Evaluations for LLM}.
The summarization of LLM evaluation is listed in Table~\ref{tbl:llm_eval}, including automatic evaluation, human evaluation, and evaluation data. 
Current LLM evaluations are primarily conducted in the form of multiple-choice questions and question-and-answers, 
which lack assessments of capabilities in clinical scenarios.

\section{Grand Challenge}
\label{sec:challenge}
Our summarization of the medical dialogue system inspires us to redesign a wide spectrum of aspects.
This section summarizes current challenges for medical dialogue systems.

\subsection{Challenges Inherited from General Domain}
\noindent
\textbf{Hallucination}.
Hallucination is defined as \textit{the generated content that is nonsensical or unfaithful to the provided source content}~\cite{filippova2020controlled,maynez2020faithfulness,zhou2020detecting}.
Hallucination in medical LLMs is concerning because it hinders performance and raises safety concerns for real-world medical applications and may lead to potential privacy violations~\cite{carlini2021extracting}.

To alleviate the issue, the popular methods are high-quality data construction, randomness reduction, retrieval-augmented generation~\cite{zhang2023mitigating,lee2022factuality}, multi-agent debate~\cite{du2023improving}, and post-process~\cite{chen2023purr,gou2023critic}.

\noindent
\textbf{Numberical Data Process}.
Medical dialogue systems often involve medical statistical data, and the understanding of the data directly affects the accuracy of system consultation.
The reason for this issue is that LLMs are probability-based generative models. 
They generate text responses from a softmax function probability distribution. 

A key solution for this issue is plug-in~\cite{schick2023toolformer,shen2023hugginggpt}, which exploits external tools to improve their capabilities.
Inspired by Toolformer~\cite{schick2023toolformer} and huggingGPT~\cite{shen2023hugginggpt}, a solution for medical numberical values is that LLMs can be designed to return mathematical expressions, perform calculations with mathematical plug-ins, and return the calculation results to the large model for reply.

\noindent
\textbf{Adversarial Attack}.
Adversarial examples are inputs designed by an adversary to cause a neural network to perform some incorrect behavior~\cite{biggio2013evasion,szegedy2013intriguing}, which may cause serious medical accidents.

The possible solutions for this issue are adversarial training~\cite{shafahi2019adversarial}, and ensemble learning~\cite{dong2020survey}.
A combination of these techniques, along with ongoing research and vigilance, can help improve the robustness of neural networks to adversarial attacks.

\subsection{Medical-specific Challenges}
\noindent
\textbf{Medical Specialization}.
Current medical LLMs can not perform as a doctor to make a clinical diagnosis and are more like a medical Q\&A.
Two examples from patient-doctor and patient-ChatGPT are shown in Figure~\ref{figure:example}.
In the example, the doctor inquires for additional patient information, provides diagnostic results, and then gives treatment advice.
However, ChatGPT lists possible diagnoses for the input question, instead of a specific conclusion.

The possible solutions for this issue are curating specialized medical training data with medical decision-making processes, and retrieval augmented generation~\cite{li2023meddm}.





\noindent
\textbf{Medical LLMs Evaluation}.
Towards the evaluation of medical LLMs’ capabilities, current evaluation methods can be divided into two main kinds, which are medical information extraction, and medical question-and-answering. 
Current LLM evaluations are insufficient for the evaluation of LLMs’ diagnostic capabilities in real clinical scenarios as they neglect either multi-turn diagnostic interviewing or rigorous diagnostic results. 
Therefore, there is a great demand for designing a unified and comprehensive evaluation criterion, such as LLM-Mini-CEX~\cite{shi2023llm} for evaluating LLMs’ diagnostic capability in real clinical applications.


\noindent
\textbf{Multi-Modal Medical Dialogue}.
Current medical dialogue systems conduct diagnoses based on text interaction.
However, a lack of multi-modal information may lead to incorrect diagnostic results.
For example, if specific images are missing during the process of seeking medical treatment for skin diseases, accurate diagnostic results can not be made for the specific disease obtained by the patient.
To alleviate the issue, multi-modal medical dialogue systems are needed to understand and process multi-modal inputs.

\noindent
\textbf{Multi-disciplinary Treatment}.
Multidisciplinary consultation provides an opportunity for specialists from different disciplines to engage in formal discussions over diagnostic and therapeutic strategies in oncology. 
In complex clinical situations, specialists discuss decisions collectively, particularly in cases involving palliative chemotherapy.

LLM-based multi-agent society is a promising method to conduct multidisciplinary consultation.
\citet{zhang2023exploring,tang2023medagents} has shown that collaborative strategies with various permutations of thinking patterns attribute significantly to performance. 

\section{Conclusion}
\label{sec:conclusion}
We summarized the progress of medical dialogue systems by introducing categories, methods before LLM, LLM-based methods, metrics, and datasets.
In addition, we discussed some new trends and their challenges, which may attract more breakthroughs.

\section*{Limitation}
This study presented a comprehensive review from a technical perspective.
However, the current version primarily focuses on technique, 
lacking analysis from a medical perspective.
In the future, we intend to include more in-depth comparative analyses to gain a better understanding of current medical dialogues from the medical perspective.

\bibliography{anthology}

\appendix
\section{Appendix}
\label{sec:appendix}

\begin{table*}[]
\small
\centering
\begin{tabular}{@{}llcccc@{}}
\toprule
Date    & Name             & Institution                                                                     & Foundation Model / Manner                                                     & \# of Parameter                                        & Code                                                        \\ \midrule
2024-01 & AMIE             & Google                                                                          & PaLM2 / Tuning                                                                 & 540B                                                   & -                                                           \\ \midrule
2023-12 & Zhongjing             & Zhengzhou University                                                                         & Ziya-LLaMA / Tuning                                                                 & 13B                                                   & \href{https://github.com/SupritYoung/Zhongjing}{Github}                                                           \\ \midrule
2023-11 & MedAgents     & Yale University                                                                   & GPT-4 / Prompting                                                            & -                                                & \href{https://github.com/gersteinlab/MedAgents}{Github}        \\ \midrule
2023-11 & MedPrompt     & Microsoft                                                                   & GPT-4 / Prompting                                                            & 7B, 13B                                                & -        \\ \midrule
2023-11 & HuatuoGPT-II     & CUHK, Shenzhen                                                                   & Baichuan2 / Tuning                                                            & 7B, 13B                                                & \href{https://github.com/FreedomIntelligence/HuatuoGPT-II}{Github}         \\ \midrule
2023-11 & MEDITRON         & EPFL                                                                            & LLaMA2 / Tuning                                                               & 70B                                                    & \href{https://huggingface.co/epfl-llm/meditron-70b}{Github}                \\  \midrule

2023-11 & Qilin-Med         & Peking University                                                                            & Baichuan / Tuning                                                               & 7B                                                    & \href{https://huggingface.co/epfl-llm/meditron-70b}{Github}                \\  \midrule

2023-08 & Radiology-LLaMA2 & University of Georgia                                                           & LLaMA2 / Tuning                                                               & -                                                      & -                                                           \\  \midrule
2023-08 & Dr. Knows & \begin{tabular}[c]{@{}c@{}}University of \\ Wisconsin Madison \end{tabular}                                                           & ChatGPT / Prompting                                                               & -                                                      & -                                                           \\  \midrule
2023-07 & CoDoC            & Google                                                                          & -                                                                     & -                                                      & \href{https://github.com/google-deepmind/codoc/tree/main}{Github}          \\ \midrule
2023-07 & CareGPT          & \begin{tabular}[c]{@{}c@{}}Macao Polytechnic \\ University\end{tabular}         & \begin{tabular}[c]{@{}c@{}}Baichuan2, LLaMA2,\\ InternLM / Tuning \end{tabular} & \begin{tabular}[c]{@{}c@{}}7B, 13B,\\ 20B \end{tabular} & \href{https://github.com/WangRongsheng/CareGPT}{Github}                    \\ \midrule
2023-05 & Med-PaLM2        & Google                                                                          & PaLM2 / Prompting                                                                & 540B                                                   & -                                                           \\ \midrule
2023-05 & Clinical Camel   & \begin{tabular}[c]{@{}c@{}}University\\ of Toronto\end{tabular}                 & LLaMA2 / Tuning                                                               & 70B                                                    & \href{https://huggingface.co/wanglab/ClinicalCamel-70B}{Hugginface}            \\ \midrule
2023-05 & DeID-GPT   & \begin{tabular}[c]{@{}c@{}}University\\ of Georgia\end{tabular}                 & GPT-4 / Prompting                                                               & -                                                    & \href{https://github.com/yhydhx/ChatGPT-API}{Github}            \\ \midrule
2023-04 & DoctorGLM        & \begin{tabular}[c]{@{}c@{}}ShanghaiTech \\ University\end{tabular}              & ChatGLM / Tuning                                                              & 6B                                                     & \href{https://github.com/xionghonglin/DoctorGLM}{Github}                   \\ \midrule
2023-04 & ChatCAD        & \begin{tabular}[c]{@{}c@{}}ShanghaiTech \\ University\end{tabular}              & ChatGPT / Prompting                                                              & -                                                     & -                   \\ \midrule
2023-04 & XrayGLM          & \begin{tabular}[c]{@{}c@{}}Macao Polytechnic \\ University\end{tabular}         & VisualGLM / Tuning                                                            & 6B                                                     & \href{https://github.com/WangRongsheng/XrayGLM?tab=readme-ov-file}{Github} \\ \midrule
2023-03 & BianQue          & \begin{tabular}[c]{@{}c@{}}South China \\ University of Technology\end{tabular} & ChatGLM / Tuning                                                              & 6B                                                     & \href{https://github.com/scutcyr/BianQue}{Github}                          \\ \midrule
2023-03 & PULSE          & \begin{tabular}[c]{@{}c@{}} Shanghai Artificial \\ Intelligence Laboratory \end{tabular}         & BLOOMZ, InternLM / Tuning                                                            & 7B, 20B                                                     & \href{https://github.com/openmedlab/PULSE}{Github} \\ \bottomrule
\end{tabular}
\caption{Some typical medical LLMs, including AMIE~\cite{tu2024towards}, Zhongjing~\cite{yang2023zhongjing}, MedAgents~\cite{tang2023medagents}, MedPrompt~\cite{nori2023can}, HuatuoGPT-II~\cite{chen2023huatuogpt}, MEDITRON~\cite{chen2023meditron}, Radionlogy-LLaMA2~\cite{liu2023radiology}, Dr. Knows~\cite{gao2023leveraging}, CoDoC~\cite{dvijotham2023enhancing}, CareGPT~\cite{wang2023caregpt}, Med-PaLM2~\cite{singhal2023towards}, Clinical Camel~\cite{toma2023clinical}, DeID-GPT~\cite{liu2023deid}, DoctorGLM~\cite{xiong2023doctorglm}, ChatCAD~\cite{wang2023chatcad}, XrayGLM~\cite{wang2023XrayGLM}, BianQue~\cite{chen2023bianque}, PULSE~\cite{pulse2023}, which are sorted by the release date of the models or the publication date of the corresponding papers and resources. ``Tuning'' and ``Prompting'' represent the fine-tuning method and the prompting method, respectively.}
\label{tbl:llm_sum}
\end{table*}

\begin{table*}[]
\small
\centering
\begin{tabular}{@{}lcccccc@{}}
\toprule
             & BM25  & DPR\_{128} & DPR\_{256} & P-DPR\_{128} & P-DPR\_{256} & Hybrid (P-DPR\_{128}) \\ \midrule
BioASQ-Small~\cite{luo2022improving} & 65.10 & 53.31  & 42.89  & 66.66    & 63.62    & 68.25             \\
BioASQ-Large~\cite{luo2022improving} & 34.32 & -      & -      & 33.13    & -        & 36.26             \\ \bottomrule
\end{tabular}
\caption{Results on medical retrieval dataset BioASQ~\cite{luo2022improving}. The evaluation metric is the mean average precision, which is expressed as percentages (\%).}
\label{tbl:retrieval_result}
\end{table*}

\begin{table*}[]
\small
\centering
\begin{tabular}{@{}llccc@{}}
\toprule
Dataset                        & Model            & Precision & Recall & F1    \\ \midrule
\multirow{6}{*}{CMDD~\cite{lin2019enhancing}}          & Bi-GRU~\cite{dyer2015transition}           & 76.02     & 88.09  & 81.61 \\
                               & Bi-LSTM~\cite{dyer2015transition}          & 76.64     & 87.60  & 81.62 \\
                               & Bi-GRU-CRF~\cite{huang2015bidirectional}       & 86.44     & 89.13  & 87.77 \\
                               & Bi-LSTM-CRF~\cite{huang2015bidirectional}      & 89.93     & 89.56  & 89.74 \\
                               & CNNs-Bi-GRU-CRF~\cite{ma2016end}  & 87.08     & \textbf{90.82}  & 88.91 \\
                               & CNNs-Bi-LSTM-CRF~\cite{ma2016end} & \textbf{90.45}     & 90.48  & \textbf{90.47} \\ \midrule
\multirow{5}{*}{MedDG~\cite{liu2022meddg}}         & LSTM~\cite{hochreiter1997long}             & 25.34     & 27.75  & 26.49 \\
                               & TextCNN~\cite{kim2014convolutional}          & 22.37     & 30.12  & 25.67 \\
                               & BERT-wwm~\cite{cui2021pre}         & 26.05     & 31.09  & 28.35 \\
                               & PCL-MedBERT~\cite{wang2022efficient}      & \textbf{26.46}     & 33.07  & 29.40 \\
                               & MedDGBERT~\cite{liu2022meddg}        & 25.34     & \textbf{36.20}  & \textbf{29.81} \\ \midrule
\multirow{7}{*}{IMCS-21 (NER)~\cite{chen2022benchmark}} & Lattice LSTM~\cite{zhang2018chinese}     & 89.37     & 90.84  & 90.10 \\
                               & BERT-CRF~\cite{devlin2018bert}         & 88.46     & 92.35  & 90.37 \\
                               & ERNIE~\cite{zhang2019ernie}            & 88.87     & 92.27  & 90.53 \\
                               & FLAT~\cite{li2020flat}             & 88.76     & 92.07  & 90.38 \\
                               & LEBERT~\cite{liu2021lexicon}           & 86.53     & \textbf{92.91}  & 89.60 \\
                               & MC-BERT~\cite{zhang2021cblue}          & 88.92     & 92.18  & 90.52 \\
                               & ERNIE-Health~\cite{zhang2019ernie}     & \textbf{89.71}     & 2.82   & \textbf{91.24} \\ \bottomrule
\end{tabular}
\caption{Results on medical information extraction datasets, including CMDD~\cite{lin2019enhancing}, MedDG~\cite{liu2022meddg}, and IMCS-21 (NER)~\cite{chen2022benchmark}. The evaluation metrics are precision, recall, and F1, which are expressed as percentages (\%).}
\label{tbl:nlu_result}
\end{table*}

\begin{table*}[]
\small
\centering
\begin{tabular}{@{}lccccccccc@{}}
\toprule
\multirow{2}{*}{Model} & \multicolumn{3}{c}{Category}                     & \multicolumn{3}{c}{Item}                         & \multicolumn{3}{c}{Full}                         \\ \cmidrule(l){2-10} 
                       & Precision      & Recall         & F1             & Precision      & Recall         & F1             & Precision      & Recall         & F1             \\ \midrule
Plain-Classifier       & 93.57          & 89.49          & 90.96          & 83.42          & 73.76          & 77.29          & 61.34          & 52.65          & 56.08          \\
MIE-Classifier-single  & 97.14          & 91.82          & 93.23          & 91.77          & 75.36          & 80.96          & 71.87          & 56.67          & 61.78          \\
MIE-Classifier-multi   & 96.61          & \textbf{92.86} & \textbf{93.45} & 90.68          & 82.41          & 84.65          & 68.86          & 62.50          & 63.99          \\
MIE-single             & 96.93          & 90.16          & 92.01          & 94.27          & 79.81          & 84.72          & 75.37          & 63.17          & 67.27          \\
MIE-multi              & \textbf{98.86} & 91.52          & 92.69          & \textbf{95.31} & \textbf{82.53} & \textbf{86.83} & \textbf{76.83} & \textbf{64.07} & \textbf{69.28} \\ \bottomrule
\end{tabular}
\caption{Result on MIE. The evaluation metrics are precision, recall, and F1, which are expressed as percentages (\%). The results are reported in category-level, item-level, and full-level.}
\label{tbl:MIE_result}
\end{table*}

\begin{table*}[]
\small
\centering
\begin{tabular}{@{}lcccc@{}}
\toprule
Models       & Precision      & Recall         & F1             & Accuracy       \\ \midrule
TextCNN~\cite{kim2014convolutional}      & 74.02          & 70.92          & 72.22          & 78.99          \\
TextRNN~\cite{liu2016recurrent}      & 73.07          & 69.88          & 70.96          & 78.53          \\
TextRCNN~\cite{lai2015recurrent}     & 73.82          & 72.53          & 72.89          & 79.40          \\
DPCNN~\cite{johnson2017deep}        & 74.30          & 69.45          & 71.28          & 78.75          \\
BERT~\cite{devlin2018bert}         & 75.35          & 77.16          & 76.14          & 81.62          \\
ERNIE~\cite{zhang2019ernie}        & \textbf{76.18} & 77.33          & 76.67          & 82.19          \\
MC-BERT~\cite{zhang2021cblue}      & 75.03          & 77.09          & 75.94          & 81.54          \\
ERNIE-Health~\cite{zhang2019ernie} & 75.81          & \textbf{77.85} & \textbf{76.71} & \textbf{82.37} \\ \bottomrule
\end{tabular}
\caption{Results of models on dialogue act classification task on IMCS-21. The evaluation metrics are precision, recall, F1, and accuracy, which are expressed as percentages (\%).}
\label{tbl:dac_result.}
\end{table*}

\begin{table*}[]
\small
\centering
\begin{tabular}{@{}llccc@{}}
\toprule
Dataset                        & Model  & Success & Match Rate & Turn \\ \midrule
\multirow{2}{*}{MZ~\cite{wei2018task}}            & DQN~\cite{liao2020task}    & 0.65    & -          & 5.11 \\
                               & KR-DQN~\cite{xu2019end} & \textbf{0.73}    & -          & -    \\ \midrule
\multirow{2}{*}{DX~\cite{xu2019end}}            & DQN~\cite{liao2020task}    & 0.731   & 0.110      & \textbf{3.92} \\
                               & KR-DQN~\cite{xu2019end} & \textbf{0.740}   & \textbf{0.267}      & 3.36 \\ \midrule
\multirow{5}{*}{IMCS-21 (DDP)~\cite{chen2022benchmark}} & DQN~\cite{liao2020task}    & 0.408   & 0.047      & 9.75 \\
                               & KR-DQN~\cite{xu2019end} & 0.485   & 0.279      & 6.75 \\
                               & REFUEL~\cite{kao2018context} & 0.505   & 0.262      & 5.50 \\
                               & GAMP~\cite{xia2020generative}   & 0.500   & 0.067      & 1.78 \\
                               & HRL~\cite{zhong2022hierarchical}    & \textbf{0.556}   & \textbf{0.295}      & \textbf{6.99} \\ \bottomrule
\end{tabular}
\caption{Results of models on medical dialogue policy learning. The evaluation metrics are success rate, match rate, and average turn. Success rate and match rate are expressed as percentages (\%).}
\label{tbl:policy_results}
\end{table*}

\begin{table*}[t]
\small
\centering
\begin{tabular}{@{}llccc@{}}
\toprule
Dataset                           & Model                                & BLEU-4                       & Distinct-1                  & Distinct-2                   \\ \midrule
                                  & Transformer~\cite{vaswani2017attention}   & 0.9   & \textbf{0.03} & 2.0   \\
                                  & BERT-GPT~\cite{zeng2020meddialog}      & 0.5   & 0.02 & \textbf{2.1}   \\
\multirow{-3}{*}{MedDialog~\cite{zeng-etal-2020-meddialog}}       & GPT-2~\cite{solaiman2019release}           & \textbf{1.8}   & 0.02 & 2.0   \\ \midrule
                                  & Seq2Seq~\cite{sutskever2014sequence}           & 19.20 & 0.75 & 5.32  \\
                                  & HRED~\cite{lei2018sequicity}          & 21.19 & 0.75 & 7.06  \\
                                  & GPT-2~\cite{solaiman2019release}         & 16.56 & \textbf{0.87} & 11.20 \\
                                  & DialoGPT~\cite{zhang2019dialogpt}      & 18.61 & 0.77 & 9.87  \\
                                  & BERT-GPT~\cite{zeng2020meddialog}      &  23.84 & 0.65 & \textbf{11.25} \\
\multirow{-6}{*}{MedDG~\cite{liu2022meddg}}           & MedDGBERT-GPT~\cite{liu2022meddg} & \textbf{23.99} & 0.63 & 11.04 \\ \midrule
                                  & Transformer~\cite{vaswani2017attention}   & 5.2   & 3.7  & 6.4   \\
                                  & GPT-2~\cite{solaiman2019release}         & \textbf{7.6}   & 13.9 & 31.0  \\
                                  & BART~\cite{lewis2019bart}          & 6.0   & \textbf{16.8} & \textbf{35.7}  \\
\multirow{-4}{*}{CovidDialog~\cite{yang2020generation}}     & BERT+TAPT~\cite{yang2020generation}     & 3.4   & 11.5 & 25.3  \\ \midrule
                                  & DialogGPT~\cite{zhang2019dialogpt}     & 0.015 & -    & -     \\
                                  & BERT~\cite{devlin2018bert}          & 0.038 & -    & -     \\
                                  & BART~\cite{lewis2019bart}          & 0.047 & -    & -     \\
\multirow{-4}{*}{Ext-CovidDialog~\cite{varshney2023knowledge}} & BioBERT~\cite{chakraborty-etal-2020-biomedbert}       & \textbf{0.048} & -    & -     \\ \midrule
                                  & BST~\cite{smith-etal-2020}           & 1.02  & -    & -     \\
                                  & MGCG~\cite{liu-etal-2020}          & 1.06  & -    & -     \\
                                  & VRbot~\cite{li2021semi}         & 1.31  & -    & -     \\
                                  & Seq2Seq~\cite{sutskever2014sequence}       & 1.01  & -    & -     \\
                                  & DialoGPT~\cite{zhang2019dialogpt}      & 1.53  & -    & -     \\
                                  & BART~\cite{lewis2019bart}          & 18.87 & -    & -     \\
\multirow{-7}{*}{MidMed~\cite{shi-etal-2023-midmed}}          & InsMed~\cite{shi-etal-2023-midmed}        & \textbf{19.73} & -    & -     \\ \bottomrule
\end{tabular}
\caption{Results of models on the medical dialogue generation task. The results are expressed as percentages (\%).}
\label{tbl:generation_result}
\end{table*}

\begin{table*}[t]
\small
\centering
\begin{tabular}{@{}lccc@{}}
\toprule
Model/Project Name & Automatic Eval                           & Human Eval   & Evaluation Data \\ \midrule
Med-PaLM~\cite{singhal2023large}            & Accuracy, Self-consistency & \cmark   & MedMCQA, PubMedQA, et al.       \\
CMB~\cite{wang2023cmb}              & GPT-4 evaluation   & \cmark  & CMB-Exam, CMB-Clin      \\
MES~\cite{tang2023evaluating}              & ROUGE-L, BLEU, etc.                    &  \cmark  &   -         \\
C-Eval (Med)~\cite{huang2023c}              &  Accuracy                                      &  \cmark  & C-Eval (Clinical Medicine, Basic Medicine)       \\
CMMLU (Med)~\cite{li2023cmmlu}              &  Accuracy                                        & \cmark  & CMMLU (College Medicine)       \\
PromptCBLUE~\cite{promptcblue}              &  Accuracy, ROUGE-L, etc.                 & \xmark                       & CMeIE, CHIP-CDEE, MedDG, et al.           \\
MedBench~\cite{MedBench}              &  Accuracy, ROUGE-L, etc.                 & \xmark                       & Med-Exam, MedHC, MedSafety, et al.            \\
\bottomrule
\end{tabular}
\caption{Summarization of LLM evaluation projects, including automatic evaluation, human evaluation, and evaluation data. Seven LLM models/ projects are listed, Med-PaLM~\cite{singhal2023large}, CMB~\cite{wang2023cmb}, MES~\cite{tang2023evaluating},
C-Eval~\cite{huang2023c}, CMMLU~\cite{li2023cmmlu}, PromptCBLUE~\cite{promptcblue}, and MedBench~\cite{MedBench}.}
\label{tbl:llm_eval}
\end{table*}

\begin{figure*}[t]
	\small
	\centering
	\includegraphics[width=\linewidth]{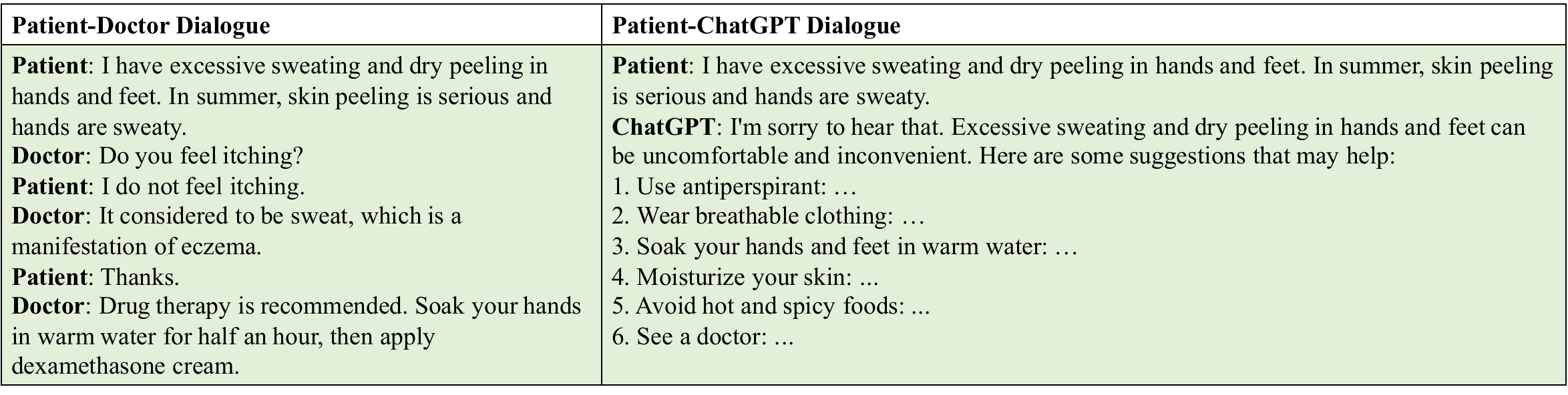}\\
	\caption{Two dialogues by patient-doctor and patient-chatgpt.}
	\label{figure:example}
\end{figure*}

\end{document}